\documentclass{Interspeech}

\interspeechcameraready 

\title{Efficient Data Selection for Domain Adaptation of ASR Using Pseudo-Labels and Multi-Stage Filtering}

\author[affiliation={1}]{Pradeep}{Rangappa} 
\author[affiliation={1}]{Andrés}{Carofilis} 
\author[affiliation={3}]{Jeena}{Prakash} 
\author[affiliation={1,2}]{Shashi}{Kumar} 
\author[affiliation={1}]{Sergio}{Burdisso} 
\author[affiliation={4}]{Srikanth}{Madikeri} 
\author[affiliation={1}]{Esaú}{Villatoro-Tello} 
\author[affiliation={3}]{Bidisha}{Sharma} 
\author[affiliation={1,5}]{Petr}{Motlicek} 
\author[affiliation={3}]{Kadri} {Hacioglu} 
\author[affiliation={3}]{Shankar} {Venkatesan} 
\author[affiliation={3}]{Saurabh}{Vyas} 
\author[affiliation={3}]{Andreas}{Stolcke} 

\affiliation{}{Idiap Research Institute , Switzerland,  $^3$Uniphore Systems, India \& USA}{}
\affiliation{}{EPFL, Switzerland,  $^4$University of Zurich, Switzerland, $^5$Brno University, Czech Republic}{}

\email{\{pradeep.rangappa, andres.carofilis, shashi.kumar,esau.villatoro,petr.motlicek\}@idiap.ch}


\usepackage{comment}
\usepackage{multirow}

\usepackage{booktabs}
\usepackage{tabularx}
\usepackage{hyperref}

\usepackage{balance}

\usepackage{pifont}

\newcommand{\WOW}{Wow }

\begin{document}

\maketitle

\begin{abstract}
Fine-tuning pretrained ASR models for specific domains is challenging for small organizations with limited labeled data and computational resources. Here we explore different data selection pipelines and propose a robust approach that improves ASR adaptation by filtering pseudo-labels generated using Whisper (encoder-decoder) and Zipformer (transducer) models. Our approach integrates multiple selection strategies---including word error rate (WER) prediction, named entity recognition (NER), and character error rate (CER) analysis---to extract high-quality training segments. We evaluate our method on Whisper and Zipformer using a 7500-hour baseline, comparing it to a CER-based approach relying on hypotheses from three ASR systems. Fine-tuning on 7500 hours of pseudo-labeled call center data achieves 12.3\% WER, while our filtering reduces the dataset to 100 hours (1.4\%) with similar performance; a similar trend is observed on Fisher English.



\end{abstract}

\section{Introduction}





Automatic speech recognition (ASR) systems have seen remarkable advancements over the past decade, transitioning from traditional hybrid models~\cite{morgan1993hybrid,bourlard1993connectionist,povey2012generating,imseng2013impact} to state-of-the-art end-to-end architectures~\cite{graves2006connectionist_ctc,gulati20_conformer,pratap2023_mms,bapna2022mslam,srinivasamurthy2017semi,viglino2019end,bhattacharjee2024minimum}. Notable models such as wav2vec 2.0, XLSR~\cite{conneau21_xlsr_53}, Conformer~\cite{gulati20_conformer,zeineldeen2022conformer}, and Zipformer~\cite{yao2024zipformer} leverage large-scale speech data to enhance transcription accuracy. Despite these advancements, real-world ASR adaptation remains constrained by the availability of in-domain annotated data. For instance, in our setting, only a few hours of annotated in-domain data are available, compared to a few hundred hours of annotated out-of-domain data and 7500 hours of unlabeled in-domain data. Given computational constraints, utilizing the entire unlabeled dataset for adaptation is impractical.

To address this challenge, recent studies have explored the use of pseudo-labeled speech data for fine-tuning ASR models~\cite{lugosch2022pseudo, peft_Geoffroy}, leveraging large quantities of unlabeled data generated in industrial pipelines. 
In real-world customer-agent interactions, only a small portion of the data is manually labeled due to the high cost of annotation.
As a result, efficient data selection techniques are crucial for leveraging pseudo-labels and fine-tuning ASR models for improved performance \cite{rangappa2025speech}.

\begin{figure}[t]
    \centering
    \includegraphics[width=\linewidth,height=5.0cm]{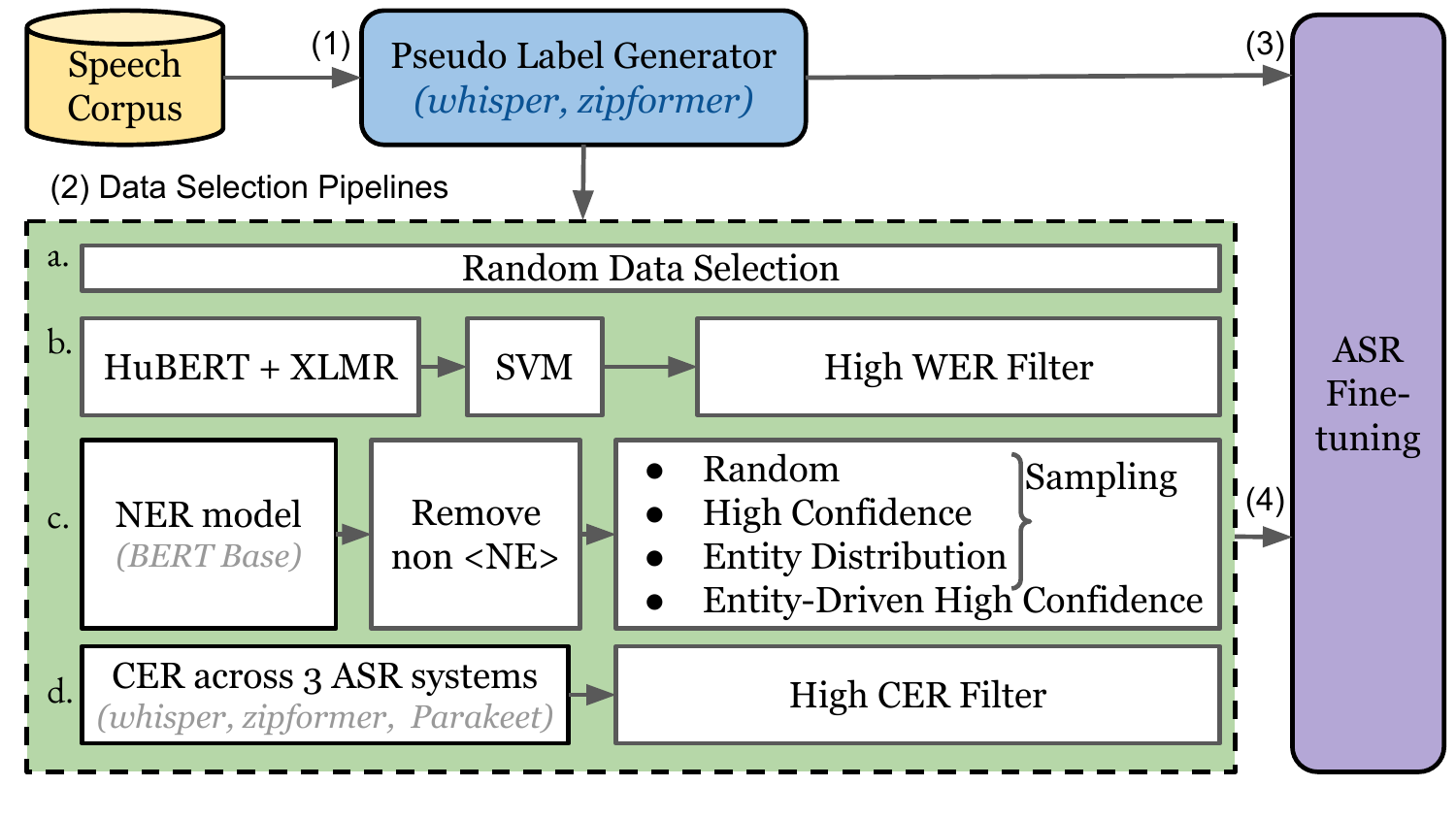}
    \caption{The functional blocks of the ASR Domain Adaptation process: (1) Pseudo-label (PL) generation using Whisper and Zipformer, where the corresponding ASR model is fine-tuned with the generated pseudo-labels. (2) Data selection pipeline consisting of: (a) Random selection, (b) WER prediction with SVM using HuBERT and XLM-R, (c) NER-based selection using a BERT model, and (d) CER-based filtering, where Nemo's Parkeet model is specifically employed along with whisper and zipformer for selecting data in the CER-based filtering step. ASR fine-tuning is carried out on the (3) full training set using the generated pseudo-labels and (4) the filtered data.}
    \label{fig:ProposedWorkOverview}
\end{figure}

We propose a robust data selection pipeline to optimize ASR adaptation by selecting high-quality training data based on three key heuristics: (1) low predicted WER, (2) high inter-ASR agreement measured via low CER, and (3) the presence of named entity (NE) classes. These heuristics are motivated by the need to mitigate error propagation, leverage the correlation between ASR agreement and transcript accuracy, and prioritize information-rich segments where NEs play a crucial role in domain-specific adaptation \cite{anoop2021unsupervised,wotherspoon2021improved}.

Our approach integrates WER classification using SVMs trained on HuBERT \cite{hsu2021hubert} and XLM-R \cite{wiciaputra2021bilingual} embeddings, CER-based filtering via majority voting across Whisper Medium, Zipformer, and Parakeet (fast conformer as encoder, RNNT as decoder) models\footnote{\url{https://huggingface.co/nvidia/parakeet-rnnt-1.1b}}, and NE-driven selection using Token2Vec and BERT-base models. 
In this work, we demonstrate that a tiny, carefully selected subset of pseudo-labeled data—comprising only 1 to 5\% of the full dataset—can match or exceed the performance achieved by fine-tuning on the entire dataset when guided by the right selection criteria.
By integrating multiple data selection strategies, we ensure computational feasibility while maximizing ASR accuracy. Our results offer empirical insights into the most effective data selection methods for ASR adaptation, providing practical guidance for future model fine-tuning efforts. Furthermore, our dynamic data selection pipeline adapts to evolving acoustic and lexical properties, optimizing both efficiency and accuracy in ASR fine-tuning. This scalable approach is particularly beneficial for real-world ASR applications, where manual annotation and computational resources are limited.

The remainder of the paper is structured as follows: Section~\ref{sec:proposed_work} describes the proposed data selection pipeline in detail. Sections~\ref{sec:experimental_setup} and~\ref{sec:results and discussion} discuss the experimental setup and results. Finally, Section~\ref{sec:conclusion} summarizes the conclusions and suggests future directions.

\section{ASR Domain Adaptation on Pseudo Labels}
\label{sec:proposed_work}

Our proposed ASR domain adaptation approach, illustrated in Figure \ref{fig:ProposedWorkOverview}, involves generating pseudo-labels using Whisper and Zipformer models. Initially, voice activity detection (VAD) is performed using PyAnnote \cite{bredin2023pyannote}, which is part of the WhisperX \cite{bain2023whisperx} framework\footnote{\url{https://github.com/SYSTRAN/faster-whisper}}, to segment the audio. For these VAD-generated segments, pseudo-labels are then extracted separately using both the Whisper and Zipformer ASR models. These pseudo-labels serve as input to the data selection pipeline, which identifies the most suitable segments for ASR finetuning. Further details are provided in Section 2.1. Finally, ASR finetuning is conducted using pretrained Whisper and Zipformer models on the entire training set, a random subset, or the selected pseudo-labeled data from our pipeline.

\subsection{Data Selection Pipelines}



We propose the following data selection pipelines for efficient ASR finetuning:
\begin{enumerate}

        \item \textbf{WER Prediction-Based Selection}: Inspired by recent studies on WER estimation \cite{park-etal-2024-automatic}, we train an SVM classifier to categorize speech segments into low- and high-WER classes. Instead of estimating WER directly, we focus on filtering out high-WER segments, prioritizing high-quality data for model finetuning. We define the boundary between low and high WER utterances at 50\%, meaning segments with WER $\leq 50\%$ are considered low-WER, while those above this threshold are categorized as high-WER. Each speech segment $s$ is represented by concatenated HuBERT-based acoustic embeddings and XLM-R-based text embeddings, denoted as $\mathbf{f}(s) = [\mathbf{f}_{\text{HuBERT}}(s); \mathbf{f}_{\text{XLM-R}}(s)]$, and classified using an SVM:  
        \[
            \hat{y}_s = \text{SVM}(\mathbf{f}(s))
        \]  
        where $\hat{y}_s \in \{ \text{low-WER}, \text{high-WER} \}$. During testing, all high-WER predicted segments are discarded, and $N$ hours of data are randomly sampled from the remaining low-WER segments:  
        \begin{equation}
            S_{\text{selected}} = \text{RandomSample}(\{ s \in S_{\text{test}} \mid \hat{y}_s = \text{low-WER} \}, N)
        \end{equation}

    \item \textbf{NER-Based Selection}: We employ a lightweight NER model, distilled from BERT \cite{devlin-etal-2019-bert}, to identify segments containing named entities.
    Specifically, we use a version of DistilBERT \cite{sanh2020distilbertdistilledversionbert} finetuned for NER,\footnote{\url{https://huggingface.co/dslim/bert-base-NER-uncased}} which achieves F1-scores of 0.99 and 0.94 for identifying samples with and without entities, respectively, on the CoNLL-2003~\cite{tjong-kim-sang-de-meulder-2003-introduction}.
    Data selection is performed in two stages: first, we identify all the segments containing entities using the NER model; then, we sample from these segments using one of the following strategies:
    \begin{itemize}
        \item \textbf{Random sampling}: Randomly select $N$ hours of data from the filtered segments containing at least one named entity class:
        \begin{equation}
            S_{\text{random}} = \text{RandomSample}(S_{\text{NER}}, N)
        \end{equation}
        where $S_{\text{NER}}$ is the set of all segments containing named entities, and $\text{RandomSample}(S, N)$ selects $N$ hours randomly.
    
        \item \textbf{High Confidence NER score based sampling}: Select the top $N$ hours of data with the highest NER confidence scores:
        \begin{equation}
            S_{\text{conf}} = \text{Top-N}(S_{\text{NER}}, N, \text{conf}(s))
        \end{equation}
        where $\text{conf}(s)$ is the confidence score assigned by the NER model to segment $s$, and $\text{Top-N}(S, N, \text{conf}(s))$ selects the $N$ most confident segments.
    
        \item \textbf{Entity Class Distribution-based Random Sampling}: Maintain entity class balance and randomly select data within each class:
        \begin{equation}
            S_{\text{dist-rand}} = \bigcup_{c \in C} \text{RandomSample}(S_{\text{NER}}^c, N_c)
        \end{equation}
        where $C$ is the set of named entity classes, $S_{\text{NER}}^c$ is the subset of $S_{\text{NER}}$ containing entity class $c$, and $N_c = P_c \cdot N$, where $P_c$ is the proportion of class $c$ in the full dataset.
    
        \item \textbf{
        Entity Class Distribution-based High Confidence Sampling}: Maintain entity class balance while selecting the most confident segments within each class:
        \begin{equation}
            S_{\text{dist-conf}} = \bigcup_{c \in C} \text{Top-N}(S_{\text{NER}}^c, N_c, \text{conf}(s))
        \end{equation}
        where $\text{Top-N}(S_{\text{NER}}^c, N_c, \text{conf}(s))$ selects the top $N_c$ hours with the highest confidence scores within class $c$.
    \end{itemize}


    \item \textbf{CER-based Selection}: 
    The unlabeled speech corpus is transcribed
    using three ASR models such as Whisper Medium, Zipformer, and Nemo Parakeet. The CER is computed for each segment across all models, and the average CER is used to make the selection. Only segments with an average CER below a predefined threshold $\tau$ (e.g., 5\%) are retained for finetuning:
    
    \begin{equation}
        S_{\text{low-CER}} = \{ s \in S_{\text{ASR}} \mid \text{CER}_{\text{avg}}(s) < \tau \}
    \end{equation} where
\begin{equation}
    \text{CER}_{\text{avg}}(s) = \frac{1}{3} \left( \text{CER}_{\text{W-Z}}(s) + \text{CER}_{\text{W-P}}(s) + \text{CER}_{\text{Z-P}}(s) \right)
\end{equation}
where $\text{CER}_{\text{W-Z}}(s)$ is the CER for segment $s$ from ASR model $whisper$ and $zipformers$.

\end{enumerate}


\section{Experiments}
\label{sec:experimental_setup}

\subsection{Datasets}\label{Data}
We evaluate our proposed approach using two datasets: the \WOW dataset, which consists of real-world call center conversations, and the Fisher English dataset, a widely used benchmark for conversational speech.
\begin{table}[h]
\centering
\caption{Overview of Datasets. The approximated total speech duration is derived from Lhotse cuts\cite{zelasko2021lhotse}.}
\resizebox{\columnwidth}{!}{%
\begin{tabular}{@{}llllll@{}}
\toprule
Dataset & Split & Transcripts & \# Conversations & \# Segments & Total Duration \\ \midrule
\multirow{3}{*}{\WOW} & train & \ding{55} & 70k & 2.58M & 7500h \\
 & dev & \checkmark & 60 & 5k & 4.5h \\
 & test & \checkmark & 250 & 26k & 18h \\
\cmidrule(r){2-6}
\multirow{3}{*}{Fisher} & train & \ding{55} & 11.5k & 1.88M & 1878h \\
 & dev & \checkmark & 25 & 5k & 3h \\
 & test & \checkmark & 23 & 5k & 3h \\ \bottomrule
\end{tabular}
}
\label{tab:dataset}
\end{table}
Below, we provide detailed descriptions of both datasets.
\begin{itemize}
    \item \textbf{\WOW}: This dataset comprises contact center conversations recorded at a sampling rate of 44.1 kHz in stereo format. The conversations span six domains: automotive, auto insurance, medicare, medical, home services, and customer services.\footnote{For more details, see: \url{https://wow-ai.com/ots-dataset.html}} Table~\ref{tab:dataset} summarizes the dataset statistics. Notably, the training set does not include ground truth transcripts, and pseudo-labels generated by the Whisper and Zipformer models are used for experimentation, reflecting the real-world scenario of limited annotations.

    \item \textbf{Fisher English}: The Fisher dataset~\cite{cieri2004fisher} is a large collection of conversational telephone speech, comprising approximately 11.5k conversations across diverse topics and speakers. The dataset is divided into training, development, and test splits, with a total duration of approximately 2,000 hours. To simulate the no-annotation scenario relevant to our study, we discard the original ground truth transcripts for the training set and use pseudo-labels generated as with \WOW for experimentation. The development and test sets retain their ground truth transcripts for evaluation purposes.
\end{itemize}
Since the ground truth WER values doesn't exists for \WOW, we used AMI \cite{kraaij2005ami} training set for training the WER classifier.

\subsection{ASR Models for Fine-tuning}

For adapting the ASR system to domain-specific data, we used the following models:

    
\begin{itemize}
    \item \textbf{Whisper Medium}: A 769M parameter model finetuned on a domain-specific dataset. The model is updated using gradient descent with a learning rate \(\eta\). Evaluation is done using task-specific metrics.

    \item \textbf{Zipformer}: A pretrained transformer-based model with approximately 70M parameters trained on Gigaspeech dataset is used to finetune the train set. It uses the ScaledAdam optimizer, a learning rate scheduler with warmup and decay, and a combined RNN-T and CTC loss. Training is done for 30 epochs with a peak learning rate of $5.0e^{-2}$ on a single RTX 3090 GPU.
\end{itemize}

\subsection{Histogram of data selected using different Data Selection Pipelines}

\begin{figure}[h]
    \centering
    \includegraphics[width=\linewidth,height=5.0cm]{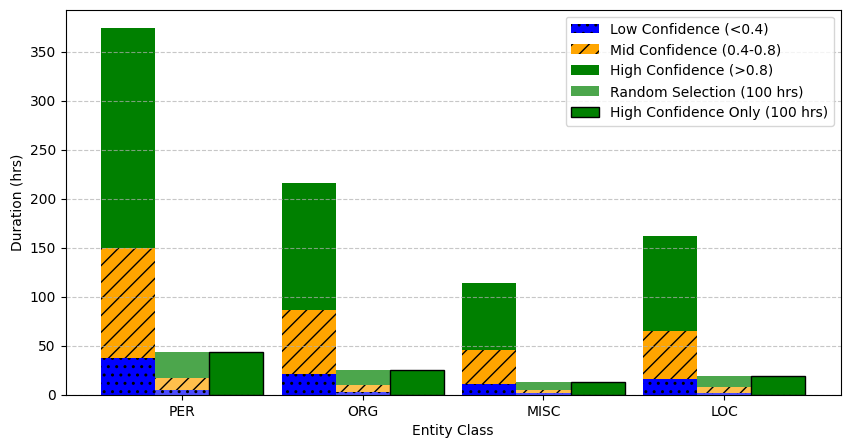}
    \caption{NER Entity Class Distribution with Confidence Levels. Each bar represents a different filtering method: Total Available Data (leftmost) shows the overall entity class distribution in grayscale, segmented by confidence levels (low, mid, high). Random Selection (100 hrs) maintains the same confidence distribution but selects segments randomly. High Confidence (100 hrs) prioritizes segments with the highest NER confidence scores ($>$0.8). Entity Class Balanced Selection (rightmost two bars) ensures proportional representation of entity classes while choosing segments either randomly or with high confidence.} 
    \label{fig:NER_DSP}
\end{figure}

Figure \ref{fig:NER_DSP} illustrates the distribution of NER-identified entity durations across four selection strategies applied to 2500 hours of the \WOW training dataset. 

We observed the average CER distribution across segments in the train set. Around 15-20\% of the total hours fall within the 0-5 CER bin, indicating a high concentration of low-error data.

\section{Results and Discussion}
\label{sec:results and discussion}

\subsection{Evaluation of the WER Estimator}

The WER classification model is evaluated on both the \WOW and AMI \cite{kraaij2005ami} test sets, with the classification results summarized in Table~\ref{tab:wer_results}.
\begin{table}[h]
    \centering
    \caption{WER Classification Results on \WOW and AMI Test Sets}
    \resizebox{\columnwidth}{!}{%
    \begin{tabular}{lcccccc}
        \toprule
        \textbf{Dataset} & \textbf{WER Class} & \textbf{Precision} & \textbf{Recall} & \textbf{F1-Score} & \textbf{Support} & \textbf{Accuracy} \\
        \midrule
        \multirow{2}{*}{AMI} & Low WER  & 0.90 & 0.95 & 0.92 & 5977 & \multirow{2}{*}{0.89} \\
                                  & High WER & 0.84 & 0.73 & 0.78 & 2340 & \\
        \cmidrule(r){2-7}
        \multirow{2}{*}{\WOW} & Low WER  & 0.85 & 0.95 & 0.89 & 18700 & \multirow{2}{*}{0.83} \\
                                  & High WER & 0.74 & 0.48 & 0.58 & 5973 & \\

        \bottomrule
    \end{tabular}%
    }
    \label{tab:wer_results}
\end{table}
We observe that the estimator performs well in distinguishing between low- and high-WER classes, achieving an overall accuracy of 83\% on the \WOW test set and 89\% on the AMI test set. The model demonstrates higher precision and recall for the low-WER class across both datasets, indicating its robustness in identifying well-recognized speech segments. 
However, the high-WER class exhibits lower recall, particularly in the \WOW test set, suggesting that high-WER utterances are often mistaken for low-WER ones.
We use competitive systems for NE detection and CER estimation as utilities to guide data selection, rather than focusing on their absolute performance metrics. Since the WOW and Fisher datasets do not contain ground-truth named entity annotations, we prioritize the utility and integration of these systems within our data selection pipeline over their standalone evaluation accuracy.


\begin{table*}[t]
\centering
\caption{WER (\%) comparison across different data selection pipelines for finetuning Whisper and Zipformer models. The table presents results for various selection criteria, including random selection, WER classification, and NER-based strategies, with performance evaluated on the two datasets: WOW and Fisher. 
For all data selection experiments, 100 hours of data were used. The baseline fine-tuning was performed on the full datasets: 7500 hours for WOW and approximately 1800 hours for Fisher.}
    \begin{tabular}{lllcccc}
        \toprule
        \textbf{Stage} & \textbf{Pipeline} & \textbf{Selection Criteria} & \multicolumn{2}{c}{\textbf{Whisper}} & \multicolumn{2}{c}{\textbf{Zipformer}} \\
        \cmidrule(lr){4-5} \cmidrule(lr){6-7}
        & & & \textbf{WOW} & \textbf{Fisher} & \textbf{WOW} & \textbf{Fisher} \\
        \midrule
        Pre-Train  & N/A      & N/A                      & 14.8 & 17.9 & 15.6 & 16.2 \\
        \midrule
        \multirow{8}{*}{Fine-tune} 
        & Baseline  & All Data                 & \textbf{12.3} & \textbf{14.8}  & \textbf{14.3} & \textbf{15.4}  \\
        \cmidrule(r){2-7} 
        & Random    & Random                   & 13.3 & 16.0 & 14.8 & 15.9 \\
        \cmidrule(r){2-7} 
        & WER Clf.  & Low WER Segments         & 13.1 & 15.1 & 14.6 & 15.5    \\
        \cmidrule(r){2-7} 
        & \multirow{4}{*}{NER} & Random                & 12.9 & 15.7 & 14.6 & 15.6 \\
        &                      & High Conf. NER Scores & 13.2 & 15.4 & 14.6 & 15.8 \\
        &                      & Entity Dist. + Random & 12.8 & 15.2 & 14.7 & 15.8 \\
        &                      & Entity Dist. + High Conf. & \textbf{12.5} & \textbf{15.0} & \textbf{14.5} & \textbf{15.5} \\
        \cmidrule(r){2-7} 
        & CER       & Segments Avg CER$<$5\%   & \textbf{12.2} & \textbf{14.1} & \textbf{13.3} & \textbf{14.6}    \\
        \bottomrule
    \end{tabular}%
\label{tab:results_asr_ft}
\end{table*}

\subsection{Baseline Performance Without Data Selection}

To establish a clear reference point, we first evaluate the performance of ASR models without any data selection strategies. This includes assessing both the performance of pretrained models on unseen data and the results of fine-tuning on the entire dataset, providing an upper bound for model performance.
\begin{itemize}
    \item \textit{No Fine-Tuning (Pre-Trained Performance)}: This is a scenario where no data is seleced as a part of finetuning. The pretrained Whisper and Zipformer models show WERs of 14.8\% and 15.6\% on WOW and 17.9\% and 16.2\% on Fisher, respectively. These results indicate that impact of pretrained models on unseen data, highlighting the need for adaptation.
    
    \item \textit{Baseline Fine-Tuning}: Here the ASR adaptation is done on all pseudo-labels of train sets (7500h for WOW and  2000h for Fisher): Fine-tuning on the full dataset significantly improves performance, bringing WERs down to 12.3\% (WOW) and 14.8\% (Fisher) for Whisper and 14.3\% (WOW) and 15.4\% (Fisher) for Zipformer. This serves as the upper bound, showing the best possible improvement with extensive labeled and pseudo-labeled data.

\end{itemize}

\subsection{Impact of Data Selection on ASR Finetuning}

Fine-tuning was performed independently for Whisper and Zipformer on both Fisher and WOW datasets. A fixed 100-hour subset was selected at the start using different data selection pipelines, ensuring consistency across experiments.

\begin{itemize}

    \item \textit{Random data selection}: With a fixed seed (42), random selection performs worse than the baseline, with WERs of 13.3\% (WOW) and 16.0\% (Fisher) for Whisper and 14.8\% (WOW) and 15.9\% (Fisher) for Zipformer. This suggests that randomly selecting 100 hours is suboptimal, as it doesn’t prioritize informative or representative samples.

    \item \textit{WER-based selection}: Selecting segments with low WER performs better than random, achieving 13.1\% (WOW) and 15.1\% (Fisher) for Whisper and 14.6\% (WOW) and 15.5\% (Fisher) for Zipformer. This approach helps in selecting cleaner, more reliable training samples, leading to modest improvements over random selection.

    \item \textit{NER-based selection}: NER-based selection outperforms WER-based selection in most cases. However, choosing only high-confidence NER segments does not guarantee better performance, sometimes performing worse than WER selection. The best NER-based method is Entity Distribution + High Confidence, which reaches 12.5\% (WOW) and 15.0\% (Fisher) for Whisper and 14.5\% (WOW) and 15.5\% (Fisher) for Zipformer—close to the baseline. This shows that a balanced selection approach is crucial rather than relying purely on high-confidence entities.

    \item \textit{CER-based selection}: Selecting segments with CER $<$ 5\% achieves the lowest WERs among all data selection strategies: 12.2\% (WOW) and 14.1\% (Fisher) for Whisper, and 13.3\% (WOW) and 14.6\% (Fisher) for Zipformer. This even surpasses the full-dataset baseline, showing that a carefully chosen subset of clean segments can be more effective than finetuning on large training sets.
    \item 
    \textit{Impact of NER/CER}: We find that the improvements come from selecting relevant data based on NER class distribution and CER confidence scores, as compared to random selection. As shown in Table~\ref{tab:results_asr_ft}, pure random and NER-filtered random subsets perform worse than CER and histogram-based NER methods; therefore, what kind of data is selected matters more than the quantity selected.

\end{itemize}

\section{Conclusions}
\label{sec:conclusion}

We have explored different data selection pipelines and proposed a robust approach that improves ASR adaptation by filtering pseudo-labels generated using Whisper and Zipformer  models. Our approach integrates multiple selection strategies---including WER prediction, NER, and character-level agreement analysis---to extract high-quality pseudo-labels.
Our experiments show that finetuning on just 1\% to 5\% of the pseudo-labeled dataset can achieve performance comparable to that of full dataset finetuning, thus offering a computationally efficient solution for small organizations with limited resources.

\section{Acknowledgments}

This work was supported by Idiap Research Institute and Uniphore collaboration project. Part of this work was also supported by EU Horizon 2020 project ELOQUENCE (grant number 101070558).

\bibliographystyle{IEEEtran}

\bibliography{mybib}

\end{document}